\documentclass[twoside]{article}

\newif\ifaccepted
\acceptedtrue

\ifaccepted
    \usepackage[accepted]{aistats2019}
\else
    \usepackage{aistats2019}
\fi
\usepackage[utf8]{inputenc} 
\usepackage[T1]{fontenc}    
\usepackage{lmodern}
\usepackage{hyperref}       
\usepackage[depth=4]{bookmark}
\usepackage{url}            
\usepackage{booktabs}       
\usepackage{amsmath}
\usepackage{amsfonts}
\usepackage{amssymb}
\usepackage{nicefrac}       
\usepackage{microtype}      
\usepackage{caption}
\usepackage{tablefootnote}
\usepackage{algorithm}
\usepackage[noend]{algorithmic}
\usepackage{stmaryrd}
\usepackage{tikz-network}
\usepackage{multirow}
\usepackage{multicol}
\usepackage{wrapfig}
\usepackage{pifont}
\usepackage{makecell}
\usepackage{titlesec}
\usepackage{array}
\usepackage{hyperref}       
\usepackage[depth=4]{bookmark}
\newcolumntype{C}[1]{>{\centering\arraybackslash}p{#1}}

\DeclareMathOperator{\diag}{diag}
\DeclareMathAlphabet\mathbfcal{OMS}{cmsy}{b}{n}
\newcommand\matrset[2]{\ensuremath{\mathbb {R} ^{#1 \times #2}}}
\newcommand{\nuovo}[1]{#1}
\newcommand{\cmark}{\ding{51}}
\newcommand{\xmark}{\ding{55}}
\newcommand{\etal}{~\emph{et al.}}
\titleformat{\paragraph}[runin]{\normalsize\bf}{}{0em}{}[.]
\newif\ifimage
\imagetrue

\begin{document}

\runningtitle{Classifying Signals on Irregular Domains via Convolutional Cluster Pooling}

\ifaccepted
    \twocolumn[
    \aistatstitle{Classifying Signals on Irregular Domains \\ via Convolutional Cluster Pooling}
    \aistatsauthor{Angelo Porrello \And Davide Abati \And Simone Calderara \And Rita Cucchiara}
    \aistatsaddress{ \\  University of Modena and Reggio Emilia, Modena, Italy \\ \texttt{name.surname@unimore.it}} ]
\else
    \twocolumn[
    \aistatstitle{Classifying Signals on Irregular Domains \\ via Convolutional Cluster Pooling}
    \aistatsauthor{Anonymous Author(s)}
    \aistatsaddress{ \vspace{0.1cm} \\  Affiliation \\ \texttt{email} } ]
\fi

\begin{abstract}
We present a novel and hierarchical approach for supervised classification of signals spanning over a fixed graph, reflecting shared properties of the dataset.
To this end, we introduce a Convolutional Cluster Pooling layer exploiting a multi-scale clustering in order to highlight, at different resolutions, locally connected regions on the input graph. 
Our proposal generalises well-established neural models such as Convolutional Neural Networks (CNNs) on irregular and complex domains, by means of the exploitation of the weight sharing property in a graph-oriented architecture. 
In this work, such property is based on the centrality of each vertex within its soft-assigned cluster. Extensive experiments on NTU RGB+D, CIFAR-10 and 20NEWS demonstrate the effectiveness of the proposed technique in capturing both local and global patterns in graph-structured data out of different domains.
\end{abstract}
\section{Introduction}
Convolutional Neural Networks (CNNs) have been successfully applied in different domains, such as speech recognition~\cite{hinton2012deep}, image classification~\cite{alex2012imagenet}, and video analysis~\cite{taylor2010convolutional}. In these domains, data can be described as a signal defined on a regular grid, whose underlying dimension can be 1d, 2d or 3d. One of the key aspects of CNNs is that such a regular structure makes it possible to exploit local and stationary properties of data. Moreover, the convolution operator, its behaviour being equivariant to translations, allows filters with a limited support on the input grid, leading to a significantly smaller number of parameters with respect to Fully Connected Networks.
However, we are surrounded by data lying on an underlying structure, which typically has an irregular and non-euclidean nature. This is the case, for instance, of document databases, 3D skeletal data, information from social networks and chemical compounds. In all these domains, the relationships among entities are more complex than in the case of a simple grid-like connectivity. Instead, graphs constitute better representation forms, because they model directly the topological structures of such data domains, through edge weights. For this reason, many efforts have recently been made~\cite{bruna2014spectral,kipf2017semi,defferrard2016cnn} in an attempt to generalise CNNs for graph-structured data.
In this work we focus on signal classification in homogeneus graphs. In such context, each sample obeys a single $\mathcal{G} = (\mathcal{V},\mathcal{E})$ weighted graph, which reflects the physics as well as the structure of the given problem. The point in which a sample differs from the others is represented by the value of each vertex in the graph. As in the case of~\cite{shuman2012emerging}, we refer to each sample as a realisation of a signal on $\mathcal{G}$. The aim is to learn a function which maps each sample into the label space. By doing so, similarly to what CNNs do for images, at each step we shall exploit information coming from the neighbouring nodes.
\ifimage
\begin{figure*}[t]
\centering
\resizebox{.85\textwidth}{!}{
\SetCoordinates[yAngle=-90, zAngle=5]
\begin{tikzpicture}[multilayer=3d]
	\tikzstyle{fake}=[draw=none,fill=none]
	\tikzstyle{fc}=[line width = 0.5pt, fill=white]
	\tikzstyle{conv}=[line width = 0.7pt]
	
	\SetLayerDistance{3.1}
	\begin{Layer}[layer=1]
		\draw[line width = 0.8pt] (-.5,-.5) rectangle (2.5,2);
	\end{Layer}
	\begin{Layer}[layer=2]
		\draw[line width = 0.8pt] (-.3,-.3) rectangle (2.3,1.8);
	\end{Layer}
	\begin{Layer}[layer=3]
		\draw[line width = 0.8pt] (+0.2,+0.2) rectangle (1.9,1.6);
	\end{Layer}
	
 	\Vertices{images/graph2/vertices.csv}
  	\Edges[vertices=images/graph2/vertices.csv,layer={1,1}]{images/graph2/edges.csv}
  	\Edges[vertices=images/graph2/vertices.csv,layer={2,2}]{images/graph2/edges.csv}
  	\Edges[vertices=images/graph2/vertices.csv,layer={1,2},style=dashed]{images/graph2/edges.csv}
  	\Edges[vertices=images/graph2/vertices.csv,layer={2,3},style=dashed]{images/graph2/edges.csv}
	
	\Vertex[x=-1.00, 	y=1.0, 	layer=4, size=.1, style=fc]{A0}
	\Vertex[x=-0.75,	y=1.0, 	layer=4, size=.1, style=fc]{A1}
	\Vertex[x=-0.50,	y=1.0,	layer=4, size=.1, style=fc]{A2}
	\Vertex[x=-0.25,	y=1.0,  layer=4, size=.1, style=fc]{A3}
	\Vertex[x=0.00,		y=1.0,	layer=4, size=.1, style=fc]{A4}
	\Vertex[x=0.25, 	y=1.0, 	layer=4, size=.1, style=fc]{A5}
	\Vertex[x=0.50,		y=1.0,	layer=4, size=.1, style=fc]{A6}
	\Vertex[x=0.75, 	y=1.0, 	layer=4, size=.1, style=fc]{A7}
	\Vertex[x=1.00,		y=1.0,	layer=4, size=.1, style=fc]{A8}
	\Vertex[x=1.25,		y=1.0,  layer=4, size=.1, style=fc]{A9}
	\Vertex[x=1.5,		y=1.0,	layer=4, size=.1, style=fc]{A10}
	\Vertex[x=1.75, 	y=1.0, 	layer=4, size=.1, style=fc]{A11}
	
	\Vertex[x=-1.25,	y=1.4,	layer=5,  size=.1, style=fc]{B1}
	\Vertex[x=-1.00,  	y=1.4,	layer=5,  size=.1, style=fc]{B2}
	\Vertex[x=-0.75,	y=1.4,	layer=5,  size=.1, style=fc]{B3}	
	\Vertex[x=-0.5,  	y=1.4,	layer=5,  size=.1, style=fc]{B4}

 \begin{scope}[on background layer]
   \foreach \alpha in {A0,A1,A2,A3,A4,A5,A6,A7,A8,A9,A10,A11}%
   {%
   \foreach \alphb in {B1,B2,B3,B4}%
   {%
   \Edge[lw=0.2pt](\alpha)(\alphb)
   }}
  
    	\Edge[lw=0.2pt,style=dashed](V25)(A0)
    	\Edge[lw=0.2pt,style=dashed](V25)(A11)
  	
   	\node[color=black] at (2.0,3.5) {$\text{CCP Layer}$};
 	\node[color=black] at (0.8,6.5) {$\text{CCP Layer}$};
 	\node[color=black] at (-1.3,11.5) {$\text{Fully Connected}$};
 	\node[color=black] at (-2.7,13.0) {$\text{Output}$};
	
 \end{scope}

\end{tikzpicture}
}
\caption{An overview of the proposed architecture. Multiple applications of the CCP layer lead to a multi-scale clustering of the input graph, exploiting both local and global properties during the information's flow from input to output. Finally, a Multi-Layer Perceptron classifies a global representation of the input signal, captured by a feature vector on a singleton graph.}
\label{fig:architecture}
\end{figure*}
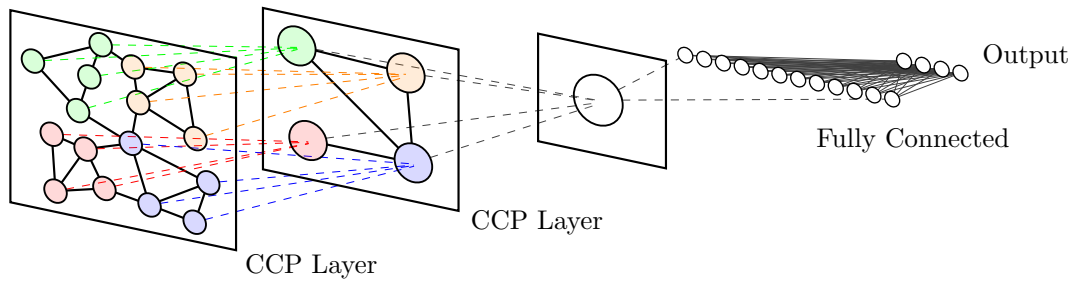
\fi
To this end, we propose a novel architecture, built by stacking multiple Convolutional Cluster Pooling (CCP) layers as depicted in Fig.~\ref{fig:architecture}. This layer, which is the main subject of this study, firstly performs a clustering operation on the input graph, resulting in a coarser output graph, whose affinity matrix reflects relationships among clusters regressed at training time. By doing so, a good basis for building local receptive fields is achieved. Secondly, according to the neighbours' vision dictated by the first step, the layer selects for each cluster a fixed number of candidate nodes for the aggregation phase, and sorts them depending on a centrality-based rank within the cluster. In this respect, it is worth noting that weight sharing across the graph's neighbourhoods can be successfully exploited.
\\
The contributions of this research are two-fold. Firstly, we provide a hierarchical framework for supervised learning in homogeneous graph contexts. Secondly, we propose a spatial formulation for graph filtering which, as for CNNs, exploits weight sharing.
\section{Related Work}
Because of its generality and potential applications in different domains, the possibility to extend neural networks to deal with graph-structured data has recently become an active research area. 
In this regard, two main approaches arise from the existing literature: spectral methods, which encode the graph structure using the graph Fourier Transform, and spatial methods, modelling the filtering operation through the construction of locally connected neighbourhoods.
\\
In general terms, spectral approaches take advantage of the fact that eigenvectors of the graph Laplacian span a space in which the convolution operator is diagonal~\cite{shuman2012emerging}. Bruna\etal~\cite{bruna2014spectral} exploited this property and defined a frequency filtering operation for neural networks. However, with such kind of formulation, it is not possible to relate the filtering operation within the spectral domain with the one performed in the vertex domain. In order to define localized linear transformations (i.e. operations also interpretable in the vertex domain~\cite{shuman2012emerging}), Defferrard\etal~\cite{defferrard2016cnn} proposed the use of polynomial spectral filters, with a theoretical guarantee of k-localisation in space. In addition, they provided a recursive approximation of such filtering through Chebyshev polynomials, which prevent expensive computations needed by the Laplacian eigenvectors.
\\\\
The other branch concerns spatial methods, which directly model convolutions as a linear combination of vertices in a local neighbourhood. In this respect, the authors of Diffusion-Convolutional Neural Networks (DCNNs)~\cite{atwood2016diffusion} presented an approach in which feature vectors are spread according to the hop distance in a depth search tree, the latter having as parent root the node for which the operation has to be done. Kipf \& Welling~\cite{kipf2017semi} proposed a fast and simple layer-wise propagation rule, which involves the use of normalized adjacency matrix. An interesting aspect of this method is how, from a spectral perspective, it may also be seen as an approximation of a localized first-order filter. Notably, the framework described by Monti\etal~\cite{monti2017geometric} led to a unified vision for all spatial approaches, in which the differences among different types of methods lie on the notion of the local coordinate system. 
\\\\
Our model is to be considered a spatial approach, because we derive a convolution-like operation directly from the clustering step, the latter creating groups of spatially close vertices itself. Inspired by Deep Locally Connected Networks~\cite{bruna2014spectral}, we then assimilate the pooling operation with the filtering stage, providing a strategy to enable weight sharing across graph's clusters.
%
Moreover, we propose a learnable multilevel strategy for graph coarsening, which may be performed directly during the learning process. On the latter point, our proposal differs from~\cite{defferrard2016cnn}, where the Graclus multilevel clustering algorithm~\cite{dhillon2007weighted} has been used, the latter being performed during a pre-processing step. On this note, we were inspired by the work of Such\etal~\cite{Such2017RobustSF}, who introduced graph embed pooling, a way to produce pooled graphs with a parametrizable number of vertices. However, our method is quite different in the computation of the pooled vertices' feature maps. Indeed, while they consider output vertices as a weighted combination of all input vertices (where weights are given by clusters' memberships), we only sample a fixed number of vertices, and combine them according to learnable kernel's weights.
Our spatial formulation builds on the concept that the weight sharing property can be inducted in a graph-oriented architecture, provided that a nodes-ordering criteria has previously been defined. A similar idea arised in PATCHY-SAN~\cite{niepert2016learning}, in which a ranking procedure and a graph normalisation technique have been used to generate local receptive fields, resulting in an adjacency matrix for each selected node. This way, the authors managed to exploit structural and local properties of input graph very well. However, the authors did not address how intermediate sub-graphs should be merged and, consequently, how that procedure should be stacked on multiple layers. The latter point could make it difficult to capture global structures with the same effectiveness. Differently, our method generates receptive fields for entire clusters, enabling graph coarsening and a hierarchical architecture.
\section{Hierarchical Graph Clustering}
A graph $\mathcal{G}$ can be defined as an ordered pair $(\mathcal{V},\mathcal{E})$, where $\mathcal{V}$ is a set of $\mathcal{N}$ nodes and $\mathcal{E} \in \mathcal{V} \times \mathcal{V}$ a set of edges. In this paper we are interested in classifying signals defined on an undirected and weighted graph, in which $\mathcal{E}$ can be described by a real symmetric matrix $\mathcal{A} \in \matrset{\mathcal{N}}{\mathcal{N}}$ which, for each couple of vertices $\mathcal{V}_i$ and $\mathcal{V}_j$ $\in$ $\mathcal{V}$, provides the strength (weight) of their connections. More generally, we refer to $\mathcal{A}$ as an affinity matrix, in which each entry $\mathcal{A}_{i,j}$ gives an affinity score between $\mathcal{V}_i$ and $\mathcal{V}_j$.
In addition to the affinity matrix, which describes the topology of the graph and the relationships between nodes, it is common practice to define a signal $\mathcal{F} : \mathcal{V} \to \mathbb {R}^{d_{IN}}$ on the vertex set, which associates a $d_{IN}$ dimensional feature vector to each node of the graph.
\ifimage

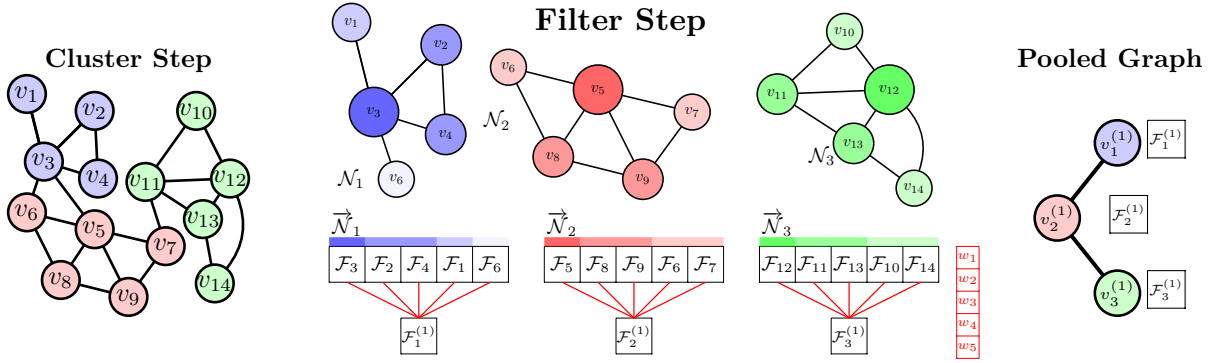
\begin{figure*}
\begin{center}
    
\begin{minipage}{0.2\textwidth}
  \begin{tikzpicture}[scale=0.45]
	\Vertices{images/graph3/vertices.csv}
	\Edges{images/graph3/edges.csv}
	\SetVertexStyle{VertexTextFont=\Huge}
	\node at (3,1){$\textbf{Cluster Step}$};
	\end{tikzpicture}
\end{minipage}
\begin{minipage}{0.05\textwidth}
\end{minipage}
\begin{minipage}{0.55\textwidth}
    \resizebox{1.0\textwidth}{!}{%
  \begin{tikzpicture}
	\Vertices{images/graph3/vertices_conv1d.csv}
	\Edges{images/graph3/edges_conv1d.csv}
	\draw[step=0.8cm,color=black] (0,-4) grid (4,-4.81);
	\draw[step=0.8cm,color=black] (4.8,-4) grid (8.8,-4.81);
	\draw[step=0.8cm,color=black] (9.6,-4) grid (13.6,-4.81);
	
	\node[scale=1.4,style={font=\footnotesize}] (A) at (0.4, -4.4) {$\mathcal{F}_3$};
	\node[scale=1.4,style={font=\footnotesize}] (B) at (1.2, -4.4) {$\mathcal{F}_2$};
	\node[scale=1.4,style={font=\footnotesize}] (C) at (2.0, -4.4) {$\mathcal{F}_4$};
	\node[scale=1.4,style={font=\footnotesize}] (D) at (2.8, -4.4) {$\mathcal{F}_1$};
	\node[scale=1.4,style={font=\footnotesize}] (E) at (3.6, -4.4) {$\mathcal{F}_6$};
	
	\node[scale=1.4,style={font=\footnotesize}] (F) at (5.2, -4.4) {$\mathcal{F}_5$};
	\node[scale=1.4,style={font=\footnotesize}] (G) at (6.0, -4.4) {$\mathcal{F}_8$};
	\node[scale=1.4,style={font=\footnotesize}] (H) at (6.8, -4.4) {$\mathcal{F}_9$};
	\node[scale=1.4,style={font=\footnotesize}] (I) at (7.6, -4.4) {$\mathcal{F}_6$};
	\node[scale=1.4,style={font=\footnotesize}] (L) at (8.4, -4.4) {$\mathcal{F}_7$};

	\node[scale=1.4,style={font=\footnotesize}] (M) at (10.0, -4.4) {$\mathcal{F}_{12}$};
	\node[scale=1.4,style={font=\footnotesize}] (N) at (10.8, -4.4) {$\mathcal{F}_{11}$};
	\node[scale=1.4,style={font=\footnotesize}] (O) at (11.6, -4.4) {$\mathcal{F}_{13}$};
	\node[scale=1.4,style={font=\footnotesize}] (P) at (12.4, -4.4) {$\mathcal{F}_{10}$};
	\node[scale=1.4,style={font=\footnotesize}] (Q) at (13.2, -4.4) {$\mathcal{F}_{14}$};
	
	\node[scale=1.4,style={font=\scriptsize}] (R) at (2.0, -6.0) {$\mathcal{F}^{(1)}_{1}$};
	\node[scale=1.4,style={font=\scriptsize}] (S) at (6.8, -6.0) {$\mathcal{F}^{(1)}_{2}$};
	\node[scale=1.4,style={font=\scriptsize}] (T) at (11.6, -6.0) {$\mathcal{F}^{(1)}_{3}$};
	
	\node[scale=1.3] at (0.4, -3.50) {$\overrightarrow{\mathcal{N}}_{1}$};
	\node[scale=1.3] at (5.2, -3.50) {$\overrightarrow{\mathcal{N}}_{2}$};
	\node[scale=1.3] at (10.0, -3.50) {$\overrightarrow{\mathcal{N}}_{3}$};
	
	\node[scale=1.3] at (0,-2.5){$\quad\quad\quad\quad$};
	\node[scale=1.3] at (0.5,-2.5){${\mathcal{N}}_{1}$};
	\node[scale=1.3] at (3.75,-1.2){${\mathcal{N}}_{2}$};
	\node[scale=1.3] at (11,-2){${\mathcal{N}}_{3}$};
	
	\draw[step=0.8cm,color=black] (1.6,-5.6) grid (2.41,-6.41);
	\draw[step=0.8cm,color=black] (6.4,-5.6) grid (7.21,-6.41);
	\draw[step=0.8cm,color=black] (11.2,-5.6) grid (12.01,-6.41);
	
	\draw [thick, - ,color=red] (0.4, -4.8) -- (2.0,-5.6);
	\draw [thick, - ,color=red] (1.2, -4.8) -- (2.0,-5.6);
	\draw [thick, - ,color=red] (2.0, -4.8) -- (2.0,-5.6);
	\draw [thick, - ,color=red] (2.8, -4.8) -- (2.0,-5.6);
	\draw [thick, - ,color=red] (3.6, -4.8) -- (2.0,-5.6);
	
	\draw [thick, - ,color=red] (5.2, -4.8) -- (6.8,-5.6);
	\draw [thick, - ,color=red] (6.0, -4.8) -- (6.8,-5.6);
	\draw [thick, - ,color=red] (6.8, -4.8) -- (6.8,-5.6);
	\draw [thick, - ,color=red] (7.6, -4.8) -- (6.8,-5.6);
	\draw [thick, - ,color=red] (8.4, -4.8) -- (6.8,-5.6);
	
	\draw [thick, - ,color=red] (10.0, -4.8) -- (11.6,-5.6);
	\draw [thick, - ,color=red] (10.8, -4.8) -- (11.6,-5.6);
	\draw [thick, - ,color=red] (11.6, -4.8) -- (11.6,-5.6);
	\draw [thick, - ,color=red] (12.4, -4.8) -- (11.6,-5.6);
	\draw [thick, - ,color=red] (13.2, -4.8) -- (11.6,-5.6);
	
	\node[scale=2.0] at (6.5,1.0){$\textbf{Filter Step}$};
	
	\draw[step=0.5cm,color=red] (13.99,-4.0) grid (14.5,-6.5);
	\node[scale=1.4,color=red,style={font=\tiny}] at (14.25, -4.25) {$w_{1}$};
	\node[scale=1.4,color=red,style={font=\tiny}] at (14.25, -4.75) {$w_{2}$};
	\node[scale=1.4,color=red,style={font=\tiny}] at (14.25, -5.25) {$w_{3}$};
	\node[scale=1.4,color=red,style={font=\tiny}] at (14.25, -5.75) {$w_{4}$};
	\node[scale=1.4,color=red,style={font=\tiny}] at (14.25, -6.25) {$w_{5}$};
	
	\draw[blue, line width=6pt, opacity=0.6] (0,-3.9) -- (0.8,-3.9);
	\draw[blue, line width=6pt, opacity=0.4] (0.8,-3.9) -- (1.6,-3.9);
	\draw[blue, line width=6pt, opacity=0.4] (1.6,-3.9) -- (2.4,-3.9);
	\draw[blue, line width=6pt, opacity=0.2] (2.4,-3.9) -- (3.2,-3.9);
	\draw[blue, line width=6pt, opacity=0.05] (3.2,-3.9) -- (4.0,-3.9);
	
	\draw[red, line width=6pt, opacity=0.6] (4.8,-3.9) -- (5.6,-3.9);
	\draw[red, line width=6pt, opacity=0.4] (5.6,-3.9) -- (6.4,-3.9);
	\draw[red, line width=6pt, opacity=0.4] (6.4,-3.9) -- (7.2,-3.9);
	\draw[red, line width=6pt, opacity=0.2] (7.2,-3.9) -- (8.0,-3.9);
	\draw[red, line width=6pt, opacity=0.2] (8.0,-3.9) -- (8.8,-3.9);
	
	\draw[green, line width=6pt, opacity=0.6] (9.6,-3.9) -- (10.4,-3.9);
	\draw[green, line width=6pt, opacity=0.4] (10.4,-3.9) -- (11.2,-3.9);
	\draw[green, line width=6pt, opacity=0.4] (11.2,-3.9) -- (12.0,-3.9);
	\draw[green, line width=6pt, opacity=0.2] (12.0,-3.9) -- (12.8,-3.9);
	\draw[green, line width=6pt, opacity=0.2] (12.8,-3.9) -- (13.6,-3.9);
	
	\end{tikzpicture}
    }%
\end{minipage}
\begin{minipage}{0.06\textwidth}
\end{minipage}
\begin{minipage}{0.12\textwidth}
  \begin{tikzpicture}
  	\Vertex[x=1.1,y=-2.3,color=blue,opacity=0.2,size=0.6,label=\tiny $v^{(1)}_1$]{A}
	\Vertex[x=0.3,y=-3.3,color=red,opacity=0.2,size=0.6,label=\tiny $v^{(1)}_2$]{B}
	\Vertex[x=1.1,y=-4.3,color=green,opacity=0.2,size=0.6,label=\tiny $v^{(1)}_3$]{C}
	\Edge[color=black](A)(B)
	\Edge[color=black](B)(C)
	
	\draw[step=0.5cm,color=black] (1.499,-2.0) grid (2.01,-2.51);
	\draw[step=0.5cm,color=black] (0.999,-3.0) grid (1.51,-3.51);
	\draw[step=0.5cm,color=black] (1.499,-4.0) grid (2.01,-4.51);
	
	\node[scale=0.8,color=black,style={font=\scriptsize}] at (1.75, -2.25) {$\mathcal{F}^{(1)}_{1}$};
	\node[scale=0.8,color=black,style={font=\scriptsize}] at (1.25, -3.25) {$\mathcal{F}^{(1)}_{2}$};
	\node[scale=0.8,color=black,style={font=\scriptsize}] at (1.75, -4.25) {$\mathcal{F}^{(1)}_{3}$};
	
	
	\node at (1,-1.2){$\textbf{Pooled Graph}$};
	\end{tikzpicture}
\end{minipage}
\caption{An illustration of the proposed CCP layer. Left, the cluster step outputs node's membership distribution among a pre-defined number of clusters. Centre, the filter step: a) selects, for each cluster, candidate nodes whose feature vectors will be aggregated; b) arranges such candidates according to a with-in cluster centrality score, building the support for the next step; c) aggregates feature vectors by means of a standard 1-d convolution, with stride equal to the kernel width (in this case, $L=5$). Right, the result of CPP layer consists in a coarsened graph coupled with its filtered pooled signal. Best viewed in color. }
\label{fig:ccp}
\end{center}
\end{figure*}
\fi
\paragraph{Graph Soft Clustering}
\label{graph_soft_clustering}
Given $\mathcal{G}=(\mathcal{V},\mathcal{E})$, we define a soft $\mathcal{K}$-partition of the graph a function that associates at each vertex $\mathcal{V}_i \in \mathcal{V}$ a membership value, in probabilistic terms, to each of the $\mathcal{|K|}$ cluster. The $\mathcal{K}$-partition can be shortly represented by a stochastic matrix $K \in \matrset{\mathcal{N}}{\mathcal{|K|}}$ where the element $K_{i,k}$ equals the probability of vertex $\mathcal{V}_i$ belonging to cluster $\mathcal{K}_k$, $\operatorname {P}(\mathcal{V}_i \in \mathcal{K}_{k})$.
Given the affinity matrix $\mathcal{A} \in \matrset{\mathcal{N}}{\mathcal{N}}$, we compute the following matrix:
\begin{equation}
\label{eq:quadratic_form}
\mathcal{A^{K}} = K^\mathrm{T} (\mathcal{A} - I_{\mathcal{N}} \odot \mathcal{A}) K,
\end{equation}
where $I_{\mathcal{N}}$ indicates the identity matrix of size $\mathcal{N}$ \footnote{The subtraction of the diagonal is performed to avoid the consideration of self-connections during cluster affinity and \textit{Cohesion} computations, in Eq. \ref{eq:affinity_eq}.}. $\mathcal{A^{K}} \in \matrset{|\mathcal{K}|}{|\mathcal{K}|}$ is highly related to the affinity matrix of the graph that can be obtained by applying to the original graph the soft $\mathcal{K}$-partition described by $K$. Indeed, if all the membership distributions behaved like a multivariate Kronecker Delta distribution, given an adjacency matrix $\mathcal{A}$ describing an undirected graph, $\mathcal{A}_{k,k}^{\mathcal{K}} \ \ k=1,2, \dots ,\mathcal{|K|}$ would be equal to the double of the number of edges existing between the nodes insides the $k$-th cluster, and $\mathcal{A}_{k,k'}^{\mathcal{K}} \ \ k,k'=1,2, \dots ,\mathcal{|K|} \ k \neq k'$ would be equal to the number of edges connecting pair of nodes respectively belonging to the $k$-th and $k'$-th cluster. Likewise, in the soft case, we have:

\begin{equation} \label{eq:affinity_eq}
     \begin{aligned}
\mathcal{A}_{k,k}^{\mathcal{K}} & = \textbf{\textit{Cohesion\,}}(K_{k})  \\
& = 2 \sum _{(\mathcal{V}_i,\mathcal{V}_j) \in\,\binom{\mathcal{V}}{2}} K_{i,k} K_{j,k} \ \mathcal{A}_{i,j}, \\
\mathcal{A}_{k,k'}^{\mathcal{K}} & = \sum _{i=1}^{\mathcal{N}}  K_{i,k} \sum _{\substack{j=1 \\ j\neq i}}^{\mathcal{N}} K_{j,k'} \ \mathcal{A}_{i,j}.
\end{aligned}
\end{equation}
In such form, $\mathcal{A}_{k,k'}^{\mathcal{K}}$ can be considered an affinity measure between the $k$-th and $k'$-th nodes in the graph reduced by $K$. We consider as a `good' soft $\mathcal{K}$-partition a partition that produces cluster with maximal cohesion. However, equivalent to ratio and normalized cut~\cite{von2007tutorial}, we penalise imbalanced solutions through the addition of a penalty related to the size of each cluster:
\begin{equation}
\begin{split}
\begin{aligned}
& \max_{K \in \mathbb {R} ^{\mathcal{N}\times \mathcal{|K|}}}
C(K) & & =\frac{1}{2} \sum _{k=1}^{\mathcal{|K|}} \frac{\textit{Cohesion\,}(K_{k})}{\textit{Vol\,}(K_{k})} \\ 
& \text{}
& & = \frac{1}{2} \mathbf{1}_{\mathcal{|K|}}^\mathrm{T} \bigg[ \diag(\mathcal{A^{K}}) \oslash (K^\mathrm{T} D) \bigg]\\
\end{aligned}
\\
\text{subject to} \quad \sum _{k=1}^{\mathcal{|K|}} K_{i,k}=1 \quad i=1,2, \dots ,\mathcal{N},\\
\end{split}
\end{equation}
where
\begin{equation}
    \textbf{\textit{Vol\,}}(K_{k}) = \sum _{i=1}^{\mathcal{N}}  D_{i} \operatorname {P} (\mathcal{V}_i \in \mathcal{K}_{k}) \quad k=1,2, \dots ,\mathcal{|K|}
\end{equation}
and $\oslash$ indicates the entry-wise division between two vectors of the same length, and $D \in \mathbb {R} ^{\mathcal{N}}$ stand for a column vector in which each entry is equal to the degree of the corresponding node. This way, we obtain clusters with maximal cohesion and, at the same time, minimum size. It is worth noting that the main difference between such formulation and the well-known normalized cut relies on the membership's definition, the latter being defined, in our case, by means of soft assignments.
\paragraph{Graph Hierarchical Soft Clustering}
Let consider $\mathcal{A}^{\mathcal{K}_{1}}$ as the affinity matrix of the graph that can be obtained by applying a soft $\mathcal{K}$-partition, given by $K^{(1)} \in \matrset{|\mathcal{K}_{0}|}{|\mathcal{K}_{1}|}$, to the original graph described by $\mathcal{A}$, where $|\mathcal{K}_{0}| = \mathcal{N}$. We can now partition $\mathcal{A}^{\mathcal{K}_{1}}$, based on the entries of a generic matrix $K^{(2)} \in \matrset{|\mathcal{K}_{1}|}{|\mathcal{K}_{2}|}$, in order to obtain a new affinity matrix $\mathcal{A}^{\mathcal{K}_{2}}$, and so on. More generally, a cascade of $\text{M}$ soft-partitions, described by an ordered sequence of $\mathcal{A}^{\mathcal{K}_{1}},\mathcal{A}^{\mathcal{K}_{2}}, \dots , \mathcal{A}^{\mathcal{K}_{\text{M}}}$, forms a soft dendrogram for the original graph. Thus, the problem of obtaining a good dendrogram, in which clusters at each level are characterized by maximal cohesion and minimum size, is formalised as follows:
\begin{equation} \label{eq:cluster_loss}
\begin{aligned}
&\max_{\substack{K^{(i)} \in \matrset{|\mathcal{K}_{i-1}|}{|\mathcal{K}_{i}|} \\ i=1,2, \dots ,\text{M}}}
& & \mathcal{L}_{\mathcal{K}} = \frac{1}{2} \sum _{m=1}^{\text{M}} \sum _{k=1}^{|\mathcal{K}_{m}|} \frac{\textit{Cohesion\,}(K_{k}^{(m)})}{\textit{Vol\,}(K_{k}^{(m)})} \\
& \text{subject to} 
& & \sum _{k=1}^{|\mathcal{K}_{m}|} K_{i,k}^{(m)}=1 {\substack{ \quad i=1,2, \dots, |\mathcal{K}_{m-1}| \\ \quad m=1,2,\dots,\text{M}. }}
\end{aligned}
\end{equation} 
\section{Convolutional Cluster Pooling}
The purpose of our proposal is to exploit the clustering mechanism in order to define a convolutional-like operator, able to ensure equivariance to translation and weight sharing in graph contexts as standard convolutions do. At a high level, our CCP operator can be considered as a layer which, at step $m$, takes in input an affinity matrix $\mathcal{A}^{\mathcal{K}_{m}}$ and a multi-dimensional $\mathcal{F}^{(m)} \in \mathbb {R} ^{|\mathcal{K}_{m}| \times d_{IN}}$ signal defined on the vertex set. The output is composed by a new reduced affinity matrix $\mathcal{A}^{\mathcal{K}_{m+1}}$ (reflecting the results of the cluster step) and a pooled signal $\mathcal{F}^{(m+1)} \in \mathbb {R} ^{|\mathcal{K}_{m+1}| \times d_{OUT}}$ (reflecting the results of the filter step where $d_{OUT}$ is the dimension of the newly computed features). All architectures used in our experiments are composed by stacking CCP layers, which combine the pooling and filtering stage and, at the same time, increase the number of feature maps, as suggested in~\cite{bruna2014spectral}. The objective in Eq.~\ref{eq:cluster_loss} is consequently optimised by backprogating gradients.
\ifimage
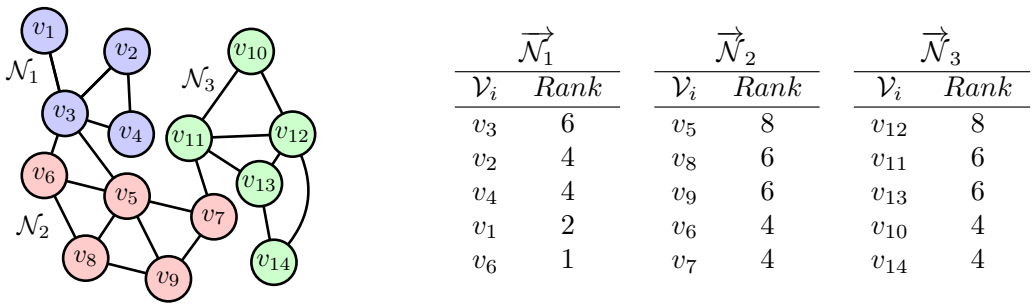
\begin{figure*}[t]
\begin{minipage}{0.40\textwidth}
\centering
\begin{tikzpicture}[scale=0.55]
	\Vertices{images/graph1/vertices_small.csv}
	\Edges{images/graph1/edges.csv}
	\node at (-.5,-1){${\mathcal{N}}_{1}$};
	\node at (-0.25,-4.75){${\mathcal{N}}_{2}$};
	\node at (3.75,-1.2){${\mathcal{N}}_{3}$};
\end{tikzpicture}
\end{minipage}
\hspace*{\fill}
\begin{minipage}{0.60\textwidth}
\resizebox{0.8\textwidth}{!}{
\begin{tabular}{ccc}
\begin{tabular}{cc}
\multicolumn{2}{c}{$\overrightarrow{{\mathcal{N}}_{1}}$}\\\hline
$\mathcal{V}_i$ & ${Rank}$\\\hline
$v_{3}$&6\\
$v_{2}$&4\\
$v_{4}$&4\\
$v_{1}$&2\\
$v_{6}$&1\\
\end{tabular}
&
\begin{tabular}{cc}
\multicolumn{2}{c}{$\overrightarrow{\mathcal{N}}_{2}$}\\\hline
$\mathcal{V}_i$ & ${Rank}$\\ \hline
$v_{5}$&8\\
$v_{8}$&6\\
$v_{9}$&6\\
$v_{6}$&4\\
$v_{7}$&4\\
\end{tabular}
&
\begin{tabular}{cc}
\multicolumn{2}{c}{$\overrightarrow{\mathcal{N}}_{3}$}\\\hline
$\mathcal{V}_i$ & ${Rank}$\\ \hline
$v_{12}$&8\\
$v_{11}$&6\\
$v_{13}$&6\\
$v_{10}$&4\\
$v_{14}$&4\\
\end{tabular}
\end{tabular}}
\end{minipage}
\caption{The ranking function (described by Equation \ref{eq:rank}) underpinning the filter step shown in Figure \ref{fig:ccp}. The node colours denote cluster memberships. All edges have weight equal to one.}
\label{fig:ranking}
\end{figure*}
\fi
\paragraph{Cluster Step}
First of all, our model performs a soft-clustering step on the input graph (Fig.~\ref{fig:ccp}, left). To the purpose we define the stochastic matrix described in Section \ref{graph_soft_clustering} as the output of a row-wise softmax applied on a variable matrix $U^{(m+1)} \in \mathbb {R} ^{|\mathcal{K}_{m}| \times |\mathcal{K}_{m+1}|}$ learned during training: 
\begin{equation} \label{eq:softmax_cluster}
K^{(m+1)}_{i,k} = P(\mathcal{V}^{(m)}_i \in \mathcal{K}^{(m+1)}_{k}) = {\dfrac {e^{U^{(m+1)}_{i,k}}}{\sum _{k'=1}^{|\mathcal{K}_{m+1}|}e^{U^{(m+1)}_{i,k'}}}}
\end{equation}
where $i = 1,2,\dots,|\mathcal{K}_{m}|$ and $k = 1,2,\dots,|\mathcal{K}_{m+1}|$. In the second place, the downsampled affinity matrix $\mathcal{A}^\mathcal{K}_{m+1}$ describing the soft-partitioned graph induced by $K^{(m+1)}$ is computed by means of the quadratic form in Eq.~\ref{eq:quadratic_form}. Eventually, we add a normalisation operation based on the degree matrix $D$~\cite{kipf2017semi} in order to prevent numerical instabilities:
\begin{equation}
\overline{\mathcal{A}}^{\mathcal{K}_{m+1}} = D^{-\frac12} \mathcal{A}^{\mathcal{K}_{m+1}} D^{-\frac12}.
\end{equation}
\paragraph{Neighbourhood selection} For each cluster $\mathcal{K}^{(m+1)}_{k}$, we select as candidate set $\mathcal{N}^{(m+1)}_{k}$ for the filtering stage the set containing the most $L$ representative nodes (where $L$ is an hyperparameter) as:
\begin{equation} \label{eq:neigh_def}
\mathcal{N}^{(m+1)}_{k}=\operatorname*{argmax}_{\mathcal{V}'\subset \mathcal{V}^{(m)}, |\mathcal{V}'|=L} \sum_{v \in \mathcal{V}'} {Rank}\ (v \shortrightarrow \mathcal{K}^{(m+1)}_{k}),
\end{equation}
where the rank of a vertex $\mathcal{V}^{(m)}_i$ for a particular cluster $\mathcal{K}^{(m+1)}_{k}$ is given by its centrality in that cluster:
\begin{equation}
\label{eq:rank}
{Rank}\ (\mathcal{V}^{(m)}_i \shortrightarrow \mathcal{K}^{(m+1)}_{k}) = (1+K^{(m+1)}_{i,k}) \sum _{\substack{j=1 \\ j\neq i}}^{|\mathcal{K}_{m}|} \mathcal{A}^{\mathcal{K}_{m}}_{i,j} K^{(m+1)}_{j,k}
\end{equation} 
More intuitively, we consider a node more central if it has a high membership value for the cluster under consideration and, at the same time, a large part of its direct neighbours nodes share the same cluster in the input graph (Fig. \ref{fig:ccp}, centre top).
\\
Further, for each cluster, we compute its features as a linear combination over the feature vectors of its inner nodes. In doing so, we want to exploit the weight sharing property across all neighbours, keeping the parameters' number under control. To this end, we create a coherent support across clusters, in terms of their inner topological structure. In this respect, our proposal is to sort candidates by their centrality within the neighbourhood and, afterwards, apply the same kernel to all clusters. 
\begin{equation}
\begin{aligned}
& \overrightarrow{\mathcal{N}}^{(m+1)}_{k} = (\mathcal{F}^{(m)}_{\phi(1)},\  \mathcal{F}^{(m)}_{\phi(2)},\  \dots, \ \mathcal{F}^{(m)}_{\phi(L)}), \\
\text{with } & \overrightarrow{\mathcal{N}}^{(m+1)}_{k}(l,i) = \mathcal{F}^{(m)}_{\phi(l),i} \substack{\quad l = 1,2,\dots,L \\ \quad i = 1,2,\dots,d_{IN}}.
\end{aligned}
\end{equation}
In simpler terms, the ordered set $\overrightarrow{\mathcal{N}}^{(m+1)}_{k}$ is recovered by sorting the candidates set $\mathcal{N}^{(m+1)}_{k}$ according to the ${Rank}$ function. By doing so, the $l$-th weight of the kernel is always multiplied by the feature vector $\mathcal{F}^{(m)}_{\phi(l)}$ being owned by the $l$-th node of the neighbour (in terms of centrality), namely $\mathcal{V}^{(m)}_{\phi(l)}$. Fig. \ref{fig:ranking} shows an example of the neighbourhood selection step for a simple graph. 
%
%
\bgroup
\newcommand{\lrarrow}[1][3pt]{\mathrel{%
   \hbox{\rule[\dimexpr\fontdimen22\textfont2-.2pt\relax]{#1}{.4pt}}%
   \mkern-4mu\hbox{\usefont{U}{lasy}{m}{n}\symbol{41}}}}
\newcolumntype{d}{p{0.01cm}}
\begin{table*}[t]
\centering
\caption{Summary of the architectures used in our experiments. We indicate with $(|\mathcal{K}|, d_{OUT})$ the number of nodes and feature maps of each layer. Note that a further softmax layer is employed to estimate class probabilities.}
\label{table:archs}
\resizebox{0.9\textwidth}{!}{
\begin{tabular}{l ccccccc l}
\textbf{Experiment} & \textbf{Input} & \multicolumn{6}{c}{\textbf{Architecture}} & \textbf{$\#$params} \\\hline\\[-0.8em]
\multirow{2}{*}{NTU RGB-D} & \multirow{2}{*}{$(2000,6)$}&
\multicolumn{5}{c}{
\multirow{2}{*}{
\begin{tabular}{ccccc}
$(512,256)$&$(128,384)$&$(32,512)$&$(8,768)$&$(1,1024)$\\
$L=16$&$L=16$&$L=8$&$L=8$&$L=8$\\
\end{tabular}}}
&\multirow{2}{*}{$\text{FC} 1024$} & \multirow{2}{*}{$\sim 14 \cdot 10^6$} \\
&&\\\hline\\[-0.8em]
\multirow{2}{*}{CIFAR-10} & \multirow{2}{*}{$(1024,3)$}&
\multicolumn{5}{c}{
\multirow{2}{*}{
\begin{tabular}{ccccc}
$(256,256)$&$(64,384)$&$(16,512)$&$(4,768)$&$(1,1024)$\\
$L=16$&$L=16$&$L=8$&$L=8$&$L=4$\\
\end{tabular}}}
&\multirow{2}{*}{$\text{FC} 1024$} & \multirow{2}{*}{$\sim 10 \cdot 10^6$} \\
&&\\\hline
\multirow{2}{*}{20NEWS} & 
\multirow{2}{*}{$(10000,1)$}&
\multicolumn{5}{c}{
\multirow{2}{*}{
\begin{tabular}{cccccc}
$(2048,128)$&$(512,192)$&$(128,256)$&$(32,384)$&$(4,512)$&$(1,512)$\\
$L=16$&$L=16$&$L=8$&$L=8$&$L=8$&$L=4$\\
\end{tabular}}}
&\multirow{2}{*}{$\text{FC} 256$} & \multirow{2}{*}{$\sim 25 \cdot 10^6$} \\
&&
\end{tabular}}
\end{table*}
\egroup
\paragraph{Neighbourhood Aggregation} The problem we face when sorting nodes by cluster centrality and then applying the same kernel to all neighbours, is that, by doing so we do not take into account the irregularity of the neighbour's shapes. 
As a matter of fact, the risk of this solution consists in the equal treatment, for different clusters, of nodes indexed in the same position by the sorting stage, whilst exhibiting considerably different centrality values.
In order to mitigate such risk, we once again use the centrality measure to implement a gating mechanism on feature vectors during the aggregation phase. The underlying idea is to make the filtering operation invariant to different neighbours and let the gating mechanism address different cluster's structures and shapes.
Roughly speaking, before applying the filtering operation described above, we are giving the centrality scores in input to a generic smoothed function $\sigma : \mathbb {R} \to (0,1)$ (e.g. the sigmoid function). Once this has been done, we perform a point-wise multiplication on the feature vectors of each candidate node. The desired effect of this operation is to attenuate information coming from distant or non-central nodes and, at the same time, preserve signals coming from nodes that reside in the inner part of the cluster. Lastly, our model computes the pooled feature vector as follows:
\begin{equation}
\mathcal{F}^{(m+1)}_{k,j} = \sum_{i=1}^{d_{IN}} \sum _{l=1}^{L} W_{l,i,j} \ (\sigma_{k,l} \cdot \overrightarrow{\mathcal{N}}^{(m+1)}_{k}(l,i) ) + b_j, 
\end{equation}
where $W \in \mathbb {R} ^{L \times d_{IN} \times d_{OUT}}$ and $b \in \mathbb{R}^{d_{OUT}}$ are learnable parameters of our CCP layer, whereas $\sigma_{k,l}$ refers to the gate's activation value computed at ${Rank}\ (\mathcal{V}^{(m)}_{\phi(l)} \rightarrow \mathcal{K}^{(m+1)}_{k})$. As shown in Fig.~\ref{fig:ccp} (centre bottom), this operation is equivalent to a 1-d convolution, enabling weight sharing across clusters.
\paragraph{Optimisation} Given a particular task, we simply add to the task-specific loss $\mathcal{L}_{0}$ (e.g. a cross-entropy) a term based on quality of the multi-level clustering solutions (Eq. \ref{eq:cluster_loss}) provided during the training phase:
\begin{equation}
\label{eq:total_loss}
\mathcal{L} = \mathcal{L}_{0} + \mathcal{L}_{\mathcal{K}}.
\end{equation}
It is important to note that the presence of the supervision signal may provide information to the process of clusters formation, backpropagating its gradient towards all $U$ variables (Eq.~\ref{eq:softmax_cluster}).
\section{Experiments}
In order to show the generality and effectiveness of our model for classification, we apply our architecture to three different domains. First, we train our model to classify human actions, given the 3D coordinates of each skeleton's joint: to this end, we evaluate it on NTU RGB-D dataset \cite{Shahroudy_2016_CVPR}. Secondly, we conduct experiments on image classification. More specifically, we use CIFAR-10 \cite{krizhevsky2009learning} as benchmark test, which is a challenging dataset for non-CNN architectures. \nuovo{Finally, we apply our solution on the 20NEWS dataset, where the goal is to address a text categorisation problem.} 
\paragraph{Implementation details}
In each experiment, all parameters are learned using Adam~\cite{kingma2014adam} as an optimisation algorithm, with an initial learning rate fixed to 0.001. We use ELU~\cite{Clevert2015FastAA} as activation function and Batch Normalization~\cite{Ioffe2015BatchNA} in all layers to speed up the convergence. Moreover, we apply dropout and $l_2$ weight regularisation (with value $10^{-4}$) to prevent overfitting, as well as standard data augmentation for CIFAR-10 and noise injection coupled with random 3d rotations for NTU RGB-D. 
All the others architectures' hyperparameters are summarised in Tab.~\ref{table:archs}. In each experiment, we subsample the input graph until its cardinality becomes equal to one: afterwards, we feed its feature vector into two fully connected layers, followed by a softmax layer providing the target class predictions.
\subsection{Classification performances}
\paragraph{Action Recognition}
The NTU RGB+D Human Activity Dataset~\cite{Shahroudy_2016_CVPR} is one of the largest datasets for human action recognition. It contains 56,880 action samples for 60 different actions, captured by the Kinect v.2 sensors. Each sample, showing a daily action performed by one or two participants, is made available in 4 different modalities: RGB videos, depth map sequences, 3D skeletal data and infrared videos. Since we are interested in graph classification, in order to perform action recognition, we just use the 3D skeletal data, represented by a temporal sequence of 25 joints\footnote{The pre-processing step on skeleton data has been performed according to the guidelines provided in~\cite{Shahroudy_2016_CVPR}.}. To this end, we model each sequence as a signal $\mathcal{F}^{(0)} \in \mathbb {R} ^{(25 \cdot T) \times 3}$ defined on a single fixed spatio temporal graph, whose structure can be summarised as follows: a vertex set $\mathcal{V}_{ST} = \{ v_{i,t} | i = 1,2,\dots,25 \ , t = 1,2,\dots,T\}$, which includes all joints captured in a fixed length sequence ($T=80$). The edge set $\mathcal{E}_{ST}$ can be defined as the union of two distinct subsets: $\mathcal{E}_{S}$, which contains all edges within each frame according to the natural human-body connectivity, and $\mathcal{E}_{T}$, which includes all edges existing between the same joint in two adjacent frame.
\begin{table}[b]
\centering
\caption{Summary of results in terms of classification accuracy for NTU RGB+D.}
\label{table:ntu_res}
\resizebox{\columnwidth}{!}{
\begin{tabular}{c c c}
\textbf{Method} & \textbf{Cross Subject} & \textbf{Cross View} \\[0.05em]\hline \\[-0.8em]
Lie Group \cite{vemulapalli2014human} & $50.1$ & $52.8$ \\
HBRNN-L \cite{yong2015hierarchical} & $59.1$ & $64.0$ \\
P-LSTM \cite{Shahroudy_2016_CVPR} & $62.9$ & $70.3$ \\
ST-LSTM+TS \cite{liu2016spatio} & $69.2$ & $77.7$ \\
TGCNN \cite{tong2018tensor} & $71.4$ & $82.9$ \\
Temporal Conv \cite{kim2017interpretable} & $74.3$ & $83.1$ \\
Deep STGC$_{K}$ \cite{li2018spatio} & $74.9$ & $86.3$ \\
C-CNN + MTLN \cite{Ke2017ANR} & $79.6$ & $84.8$ \\
\textbf{CCP (our)} & $\mathbf{80.1}$ & $\mathbf{86.8}$ \\
\end{tabular}}
\end{table}
In order to evaluate the model's performance, as described in~\cite{Shahroudy_2016_CVPR}, we run two different standard benchmarks: the cross-subject setting, in which the train/test split is based on two disjoint sets of actors; and the cross-view setting, where the test samples are captured from a different camera from those collecting the training sequences. We report in Tab.~\ref{table:ntu_res} the top-1 classification accuracy on both settings, comparing it with other existing approaches. As illustrated, CCP outperforms previous state-of-the-art methods on this dataset, including other graph-oriented architectures~\cite{tong2018tensor,li2018spatio}, despite being more general and not specifically designed to only address action recognition settings. 
\begin{figure}[b]
\centering
\setlength{\tabcolsep}{0.2em} 
\begin{tabular}{cccc}
1st CCP& 2nd CCP& 3rd CCP& 4th CCP \\
\includegraphics[width=0.22\columnwidth]{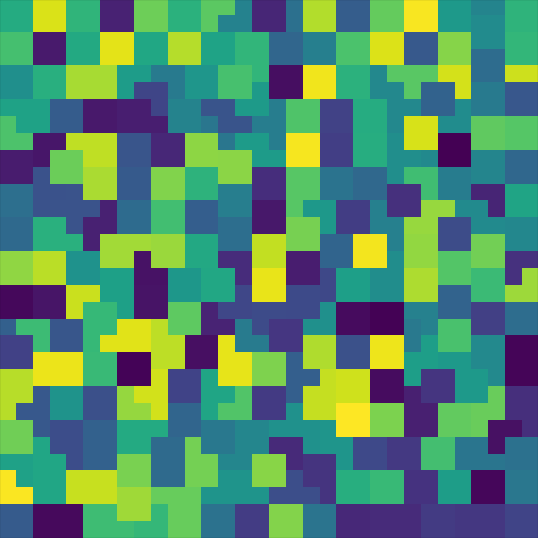}&
\includegraphics[width=0.22\columnwidth]{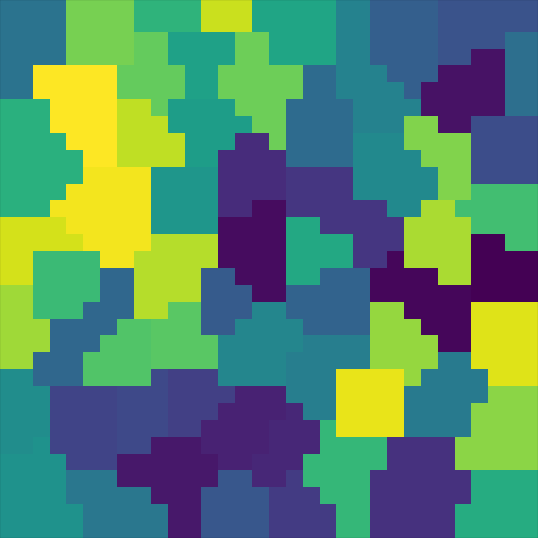}&
\includegraphics[width=0.22\columnwidth]{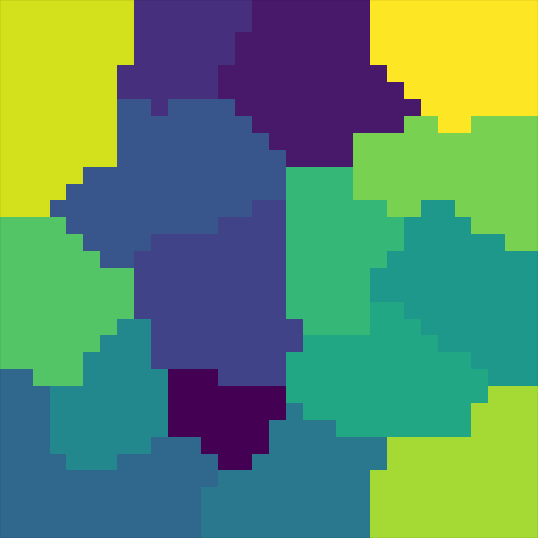}&
\includegraphics[width=0.22\columnwidth]{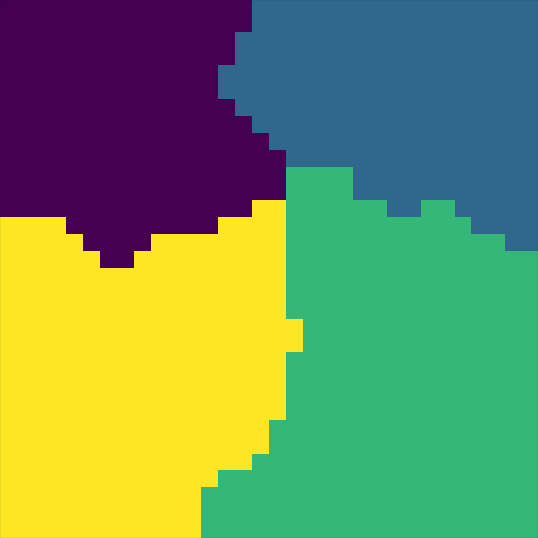}\\
\end{tabular}
\captionof{figure}{Receptive fields learned during CIFAR10 training.}
\label{fig:rec_fields}
\end{figure}
\paragraph{Image Classification}
We conduct experiments on CIFAR-10, a popular dataset widely used for image recognition. 
Each image, labeled into one of ten classes, can be treated as a signal defined on a graph, which can in turn be modeled as a 32$\times$32 grid structure. 
In particular, every pixel is a vertex such a graph, linked to its neighbours following a 8-connectivity. The colour information is encoded as a signal $\mathcal{F}^{(0)} \in \mathbb {R} ^{1024 \times 3}$ over such vertexes.
%
\begin{table}[t]
\centering
\caption{Image classification accuracy on CIFAR-10.}
\label{table:cifar_res}
\begin{tabular}{c c}
\textbf{Method} & \textbf{Accuracy} \\[0.05em]\hline \\[-0.8em]
Graph-CNNs \cite{Such2017RobustSF} & $68.3$ \\
FC \cite{Lin2015HowFC} & $78.6$ \\
\textbf{CCP (our)} & $\mathbf{84.4}$ \\
Stochastic Pooling \cite{zeiler2013stochastic} & $84.9$ \\
ResNet \cite{he2016deep} & $93.6$ \\
\end{tabular}
\end{table}
As shown in Tab.~\ref{table:cifar_res}, CCP obtains an encouraging performance in terms of classification accuracy on test set. Indeed, our method outperforms both the best reported fully connected (FC) network~\cite{Lin2015HowFC} and Graph-CNNs~\cite{Such2017RobustSF} - to the best of our knowledge, the only graph classification model in literature that reports results on CIFAR-10 - by a significant margin. To put our results into perspective, we report the performance obtained by~\cite{zeiler2013stochastic}, which is the nearest score founded in the literature given by a deep CNN, as well as the results of a state-of-art CNNs like~\cite{he2016deep}. The gap with respect to the latter is still consistent, suggesting that there is still room for improvement in euclidean domains. 
Fig.~\ref{fig:rec_fields} also depicts an illustration of the hierarchical clustering computed on the input grid. As can be seen, as the input image undergoes CCP layers, its representations are computed out of compact regions, resembling dyadic clustering that has been proven a successful downsampling strategy in CNNs.
\paragraph{Text Categorisation}
\nuovo{In order to further validate the quality of our proposal in diverse data domains, we apply our model on text categorisation. In this respect, we conduct experiments on the 20NEWS dataset~\cite{joachims1996probabilistic}, adhering to the guidelines described in~\cite{defferrard2016cnn} for the construction of the shared graph. 
To summarise, such protocol models each text as a graph which has a node for each common word in the document set. On the other hand, the pairwise connectivities of such graph are shared and obtained assessing the similarities inducted by word2vec embeddings~\cite{mikolov2013efficient}, followed by a discretisation step computed through a $K$-NN pass (with $K=16$). This way, each document $\mathcal{D}$ can be represented as a signal over a fixed graph, implemented as the word's distribution observed in $\mathcal{D}$.}
\begin{table}[t]
\centering
\caption{Text categorisation accuracy on 20NEWS.}
\label{table:20news_res}
\begin{tabular}{c c}
\textbf{Method} & \textbf{Accuracy} \\[0.05em]\hline \\[-0.8em]
Linear SVM $\dagger$  & $65.9$ \\
Softmax $\dagger$  & $66.3$ \\
Multinomial Naive Bayes $\dagger$  & $68.5$ \\
FC2500-FC500 $\dagger$  & $65.8$ \\
Chebyshev - GC32 \cite{defferrard2016cnn} $\dagger$  & $68.3$ \\
\textbf{CCP (our)} & $\mathbf{70.1}$ \\
\multicolumn{2}{c}{$\dagger$ \footnotesize{\nuovo{Baselines' results published in \cite{defferrard2016cnn}.}}}
\end{tabular}
\end{table}
\nuovo{As indicated in Tab.~\ref{table:20news_res}, the discussed approach leads to good performances, defeating both baselines and the graph convolutional layer based on polynomial spectral filters. On this latter point, our architecture seems to take advantages of its depth and hierarchical nature, differently from \cite{defferrard2016cnn} where a shallow graph convolutional network has been employed to categorise documents.} 
\subsection{Model analysis}
\paragraph{Comparisons with other coarsening approaches}
\begin{table}[b]
\centering
\caption{Comparison of different graph corsening and filtering approches on CIFAR-10 and Cross Subject NTU RGB+D.}
\label{table:competitors}
\resizebox{\columnwidth}{!}{
\begin{tabular}{l c c c c}
\Xhline{2\arrayrulewidth}
\multicolumn{1}{c}{\textbf{Filter}} & \textbf{Coarsen} & \textbf{GAP} &\textbf{CIFAR} & \textbf{NTU-CS} \\ [0.05em]\hline \\[-0.8em]
Chebyshev \cite{defferrard2016cnn} & Graclus & \xmark &$78.15$ & $74.85$ \\
GCN \cite{kipf2017semi} & Graclus & \xmark & $67.01$ & $62.00$ \\
GAT & Graclus & \xmark & $72.82$ & $59.48$\\
GAT & - & \cmark & $66.39$ & $26.74$ \tablefootnote{We found it extremely hard to train, due to the huge memory footprint required, even for a shallow configuration.}  \\[0.05em]\hline \\[-0.8em]
\textbf{CCP (ours)} & CCP & \xmark & $\mathbf{84.4}$  & $\mathbf{80.1}$\\
\Xhline{2\arrayrulewidth}
\end{tabular}}
\end{table}
\nuovo{We further compare our proposal w.r.t. three different works (Tab.~\ref{table:competitors}): GCN~\cite{kipf2017semi}, Chebyshev filtering~\cite{defferrard2016cnn} and Graph Attention Networks (GAT)~\cite{velickovic2018graph}. In this respect we use the Graclus algorithm~\cite{dhillon2007weighted} for coarsening the input graph and vary the graph filtering strategy accordingly to the referenced work. For all the experiments we keep architectural settings described in Tab.~\ref{table:archs} and use the public implementation of these works. Futhermore, we also design a non-coarsening baseline by performing a global average pooling (GAP) on nodes features (after GAT manipulation) before the fully connected classification layers. The experiment suggests that CCP outperforms, by a consistent margin, the previous GCN+coarsening approach. Moreover, it still outperforms Graclus as a coarsening strategy even though recent filters such as GAT are applied. In this regard, we empirically observed that order-invariant filters (e.g. GAT), despite being more general, may treat the same graph differently, according to the attention scores. This is in fact a great advantage when graph layout may vary across examples, though potentially unrewarding when the support remains the same through all the dataset.}
\begin{table}[tb]
\caption{Ablative results under different optimisations.}
\label{table:abl_1}
\centering
\resizebox{\columnwidth}{!}{
\begin{tabular}{c c c c}
\Xhline{2\arrayrulewidth}
\textbf{Loss $\mathcal{L}$} & \textbf{Gradient} &\textbf{CIFAR-10} & \textbf{NTU-CS} \\[0.05em]\hline \\[-0.8em]
$\mathcal{L}_{0} + \mathcal{L}_{\mathcal{K}}$ (Eq.~\ref{eq:total_loss})&-& $\mathbf{84.4}$ & $\mathbf{80.1}$\\
$\mathcal{L}_{0}$ &-& 66.7 & 73.1 \\
$\mathcal{L}_{0}$ &$\delta \mathcal{L}_{0}/\delta U \leftarrow 0$ & 56.8 & 70.6 \\
$\mathcal{L}_{0} + \mathcal{L}_{\mathcal{K}}$ & $\delta \mathcal{L}_{0}/\delta U \leftarrow 0$ & 83.8 & 78.6\\
\Xhline{2\arrayrulewidth}
\end{tabular}}
\end{table}
\begin{figure}[t]
\centering
\bgroup
\setlength{\tabcolsep}{.16667em}
\resizebox{\columnwidth}{!}{
\begin{tabular}{cc@{\hskip 0.05in}|@{\hskip 0.05in}cc}
\textbf{\huge{CCP 1}}  & \textbf{\huge{CCP 1}} & \textbf{\huge{CCP 3}} & \textbf{\huge{CCP 3}}\\
\includegraphics[width=0.3\textwidth]{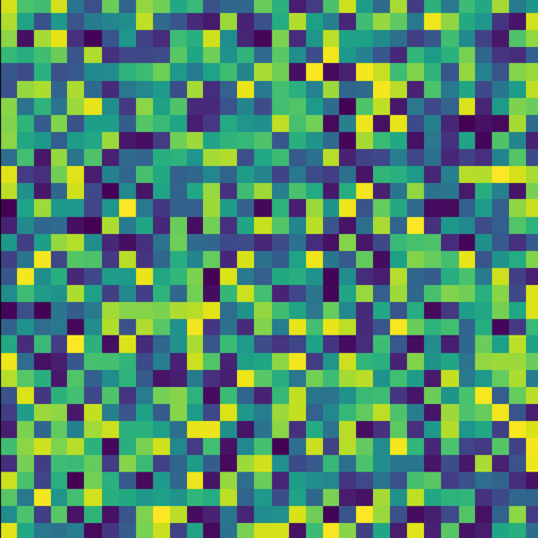}& \includegraphics[width=0.3\textwidth]{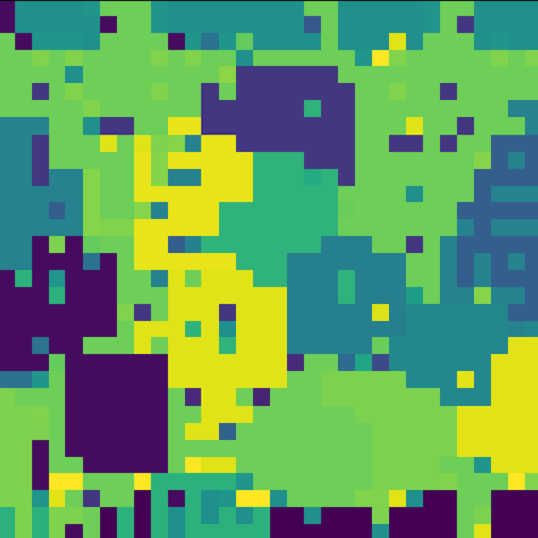}&
\includegraphics[width=0.3\textwidth]{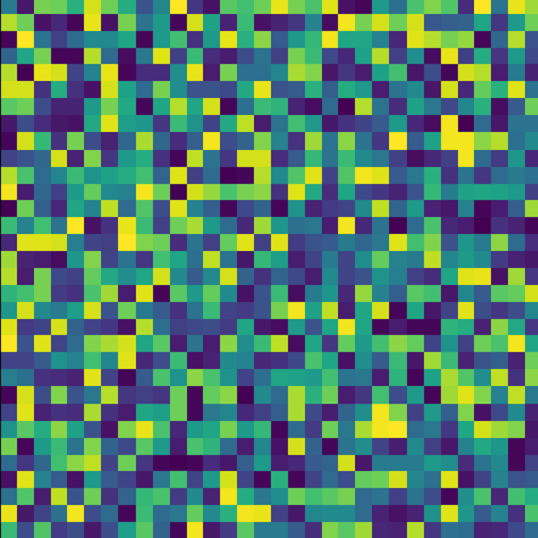}& \includegraphics[width=0.3\textwidth]{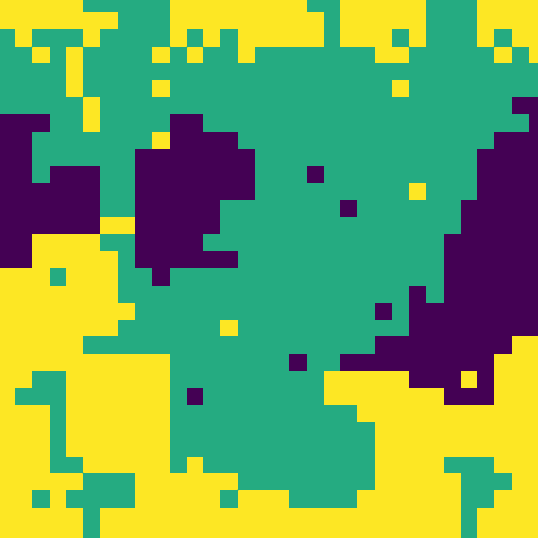}\\
\textbf{\huge{Epoch 0}} & \textbf{\huge{Epoch 500}} & \textbf{\huge{Epoch 0}} & \textbf{\huge{Epoch 500}}\\
\end{tabular}}
\captionof{figure}{Receptive fields arising from $\mathcal{L}_{0}$ minimisation on CIFAR-10.}
\label{fig:rec_abl}
\egroup
\end{figure}
\begin{figure}[b]
\centering
\includegraphics[width=0.7\columnwidth]{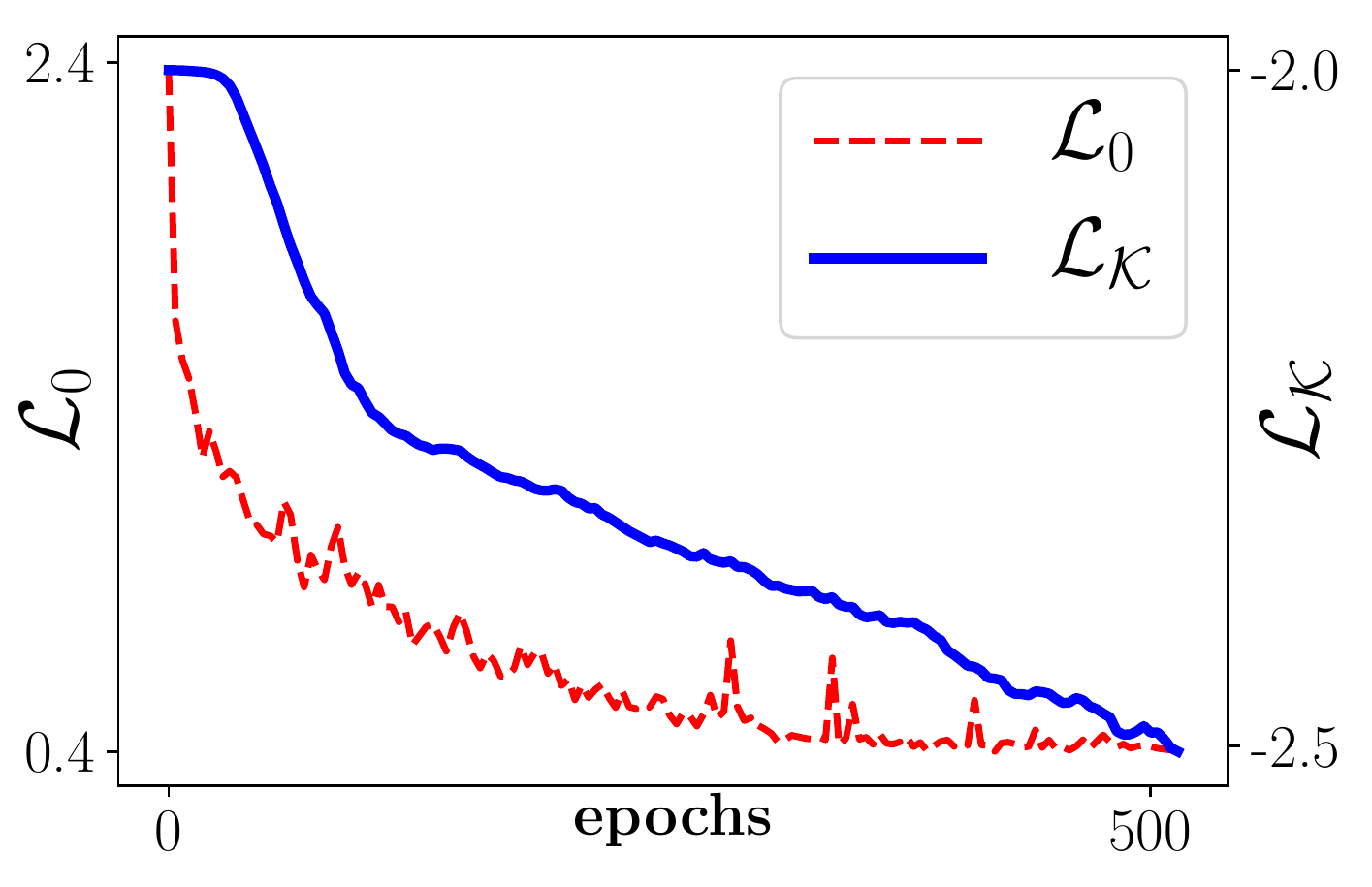}
\caption{Loss landscapes under $\mathcal{L}_{0}$ minimisation.} 
\label{fig:ce_vs_tl}
\end{figure}
\paragraph{The impact of the task-specific loss} \nuovo{We studied the contribution of the loss $\mathcal{L}_{0}$ (Eq.~\ref{eq:total_loss}) and found three evidences supporting its beneficial effect: 
i)~if only $\mathcal{L}_{0}$ is optimised, then suppressing its gradients on membership variables (i.e. cluster memberships are randomly fixed and cannot be changed during training) leads to poorer performances ($\mathcal{L}_{0}, \delta \mathcal{L}_{0}/\delta U \leftarrow 0$ against $\mathcal{L}_{0}$, Tab.~\ref{table:abl_1}); ii)~when both objectives are optimised, discarding gradients of $\mathcal{L}_{0}$ on membership variables yields slightly degraded results ($\mathcal{L}_{0} + \mathcal{L}_{\mathcal{K}}, \delta \mathcal{L}_{0}/ \delta U \leftarrow 0$, against $\mathcal{L}_{0} + \mathcal{L}_{\mathcal{K}}$, Tab.~\ref{table:abl_1}); iii)~even when only optimising $\mathcal{L}_{0}$ we can observe the emergence of compact regions (i.e. clusters) in the clustering landscape (Fig.~\ref{fig:rec_abl}). Another evidence of this effect is the lowering of the $\mathcal{L}_{\mathcal{K}}$ when the network is optimised w.r.t. $\mathcal{L}_{0}$ (Fig.~\ref{fig:ce_vs_tl}).}
\paragraph{Effectiveness of the Ranking function} Finally, we conducted an ablation study for validating the effectiveness of the proposed within-cluster centrality measure in capturing shift-invariant structures on the graph. To this end, we compared it with a less principled criteria, involving a random permutation of the candidate nodes.
Specifically, under the random setting, we still keep the definition of neighbourhood $\mathcal{N}^{(m+1)}_{k}$ given by Eq.~\ref{eq:neigh_def}. 
However, instead of sorting nodes inside it, we randomly sample a fixed permutation for each cluster, before the start of the learning. Without the sorting criteria, learnable kernels cannot rely on a coherent topological structure within their support, weakening the effect of weight sharing. We evaluated both of the policies on CIFAR-10, and reported our results in Fig.~\ref{table:abl}, in terms of learning curves and test error. The figure suggests that the sorting criterion indeed leads to a significant improvement in performance, due to a proper exploitation of weight sharing.
\begin{figure}[t]
\centering
\includegraphics[width=\columnwidth]{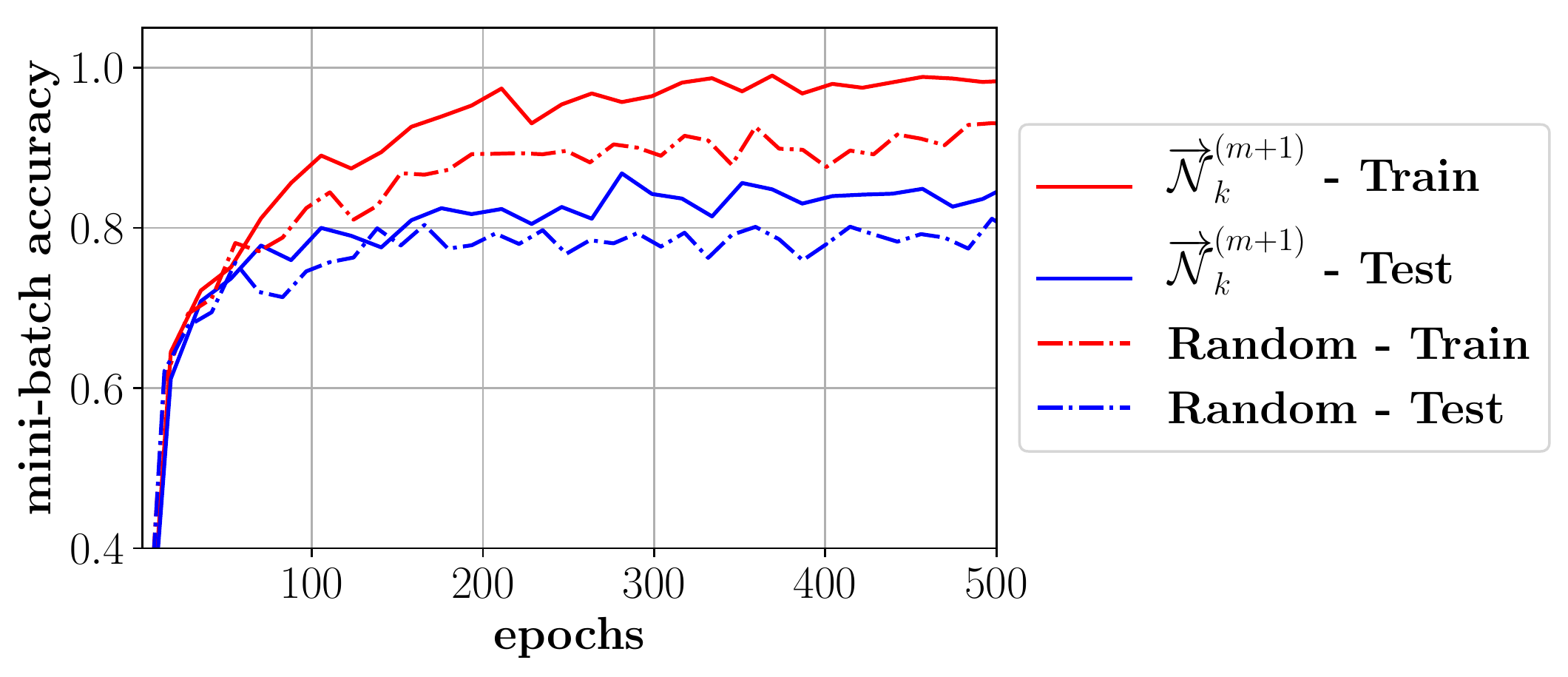}\\
\begin{tabular}{cc}
\textbf{Order} & \textbf{Accuracy} \\ [0.05em]\hline \\[-0.8em]
Cluster Centrality & $84.4$ \\
Random & $80.3$ \\
\end{tabular}
\caption{Results from the ablation study conducted on CIFAR-10. The top picture shows test and training learning curve under both settings.}
\label{table:abl}
\end{figure}
\section{Conclusion}
In this paper, we have proposed a novel approach for graph signal classification, leveraging both local and global structures, the latters arising from a multi-scale and hierarchical representation of the input signal. Our main contribution consists in a layer which performs a (soft) clustering step on the input graph and, accordingly, aggregates information within each cluster. Experiments show that our model consistently outperforms recent graph-based classification models in different data domains. The ablation study suggests that the proposed layer successfully exploit the weight sharing property in a graph convolutional architecture. For future works, we aim to generalise our architecture for vertex classification tasks and heterogeneous graphs, as well as further investigate the impact of supervision during clusters formation, which could lead to task-dependent pooling regions.
\subsubsection*{Acknowledgments} The authors thank Ottavia Credi for the assistance she provided with the editing and revision of the paper. This work was done within the UNIMORE project interdisciplinary FAR 2016 – UBINVB – Ubiquitous objective measures of intergroup nonverbal behaviors.
%
\bibliographystyle{plain}

\clearpage
\section{Supplementary Material}
\begin{figure*}[h!]
  \centering
  \begin{minipage}{.7\linewidth}
    \begin{algorithm}[H]
   \caption{Cluster Step}
   \label{alg:ga1}
\begin{algorithmic}
   \STATE {\bfseries Input:} affinity matrix $\mathcal{A}^{\mathcal{K}_{m}}$, number of output clusters $|\mathcal{K}_{m+1}|$
   \STATE {\bfseries Output:} affinity matrix $\mathcal{A}^{\mathcal{K}_{m+1}}$
   \IF{$\textbf{init}$}
   \STATE $U^{(m+1)} \leftarrow \text{random}(|\mathcal{K}_{m}|,|\mathcal{K}_{m+1}|)$
   \ENDIF
   \STATE $K^{(m+1)} \leftarrow \text{rowsoftmax}(U^{(m+1)})$
   \STATE ${\mathcal{A}}^{\mathcal{K}_{m+1}} \leftarrow K^{(m+1)} {}^\mathrm{T} (\mathcal{A}^{\mathcal{K}_{m}} - I_{\mathcal{N}} \odot \mathcal{A}^{\mathcal{K}_{m}}) K^{(m+1)}$ 
   \STATE $D \leftarrow {\mathcal{A}}^{\mathcal{K}_{m+1}} \mathbf{1}_{|\mathcal{K}_{m+1}|} $ 
   \STATE $\overline{\mathcal{A}}^{\mathcal{K}_{m+1}} \leftarrow D^{-\frac12} {\mathcal{A}}^{\mathcal{K}_{m+1}} D^{-\frac12}$ 
   \STATE {\bfseries Return: $\overline{\mathcal{A}}^{\mathcal{K}_{m+1}}$} 
\end{algorithmic}
\end{algorithm}
\end{minipage}
\end{figure*}
\subsection{Pseudocode}
This section introduces algorithms involved in the proposed CCP layer discussed in Section 4 of the paper. Specifically, Algorithm~\ref{alg:ga1} illustrates the cluster step (Figure 2, left in the main paper), whereas Algorithm~\ref{alg:ga2} details the filter step (Figure 2, center in the main paper)
\begin{figure*}[ht]
  \centering
  \begin{minipage}{.7\linewidth}
    \begin{algorithm}[H]
   \caption{Filter Step}
   \label{alg:ga2}
    \begin{algorithmic}
    \STATE {\bfseries Input:} normalized affinity matrix $\overline{\mathcal{A}}^{\mathcal{K}_{m}}$, input feature maps $\mathcal{F}^{(m)} \in \mathbb {R} ^{|\mathcal{K}_{m}| \times d_{IN}}$ , normalized and reduced affinity matrix $\overline{\mathcal{A}}^{\mathcal{K}_{m+1}}$, number of output channels $d_{OUT}$, filter size $L$ 
   \STATE {\bfseries Output:}, output feature maps $\mathcal{F}^{(m+1)} \in \mathbb {R} ^{|\mathcal{K}_{m+1}| \times d_{OUT}}$
   \STATE
   \IF{$\textbf{init}$}
   \STATE $W,b \leftarrow random(L,d_{IN},d_{OUT}),\ random(C_{OUT})$
   \STATE $\alpha,\beta \leftarrow \alpha \sim \mathcal{N}(\mu = 1,\,\sigma_{1}^{2}),\ \beta \sim \mathcal{N}(\mu = 0,\,\sigma_{2}^{2}) $
   \ENDIF
   \STATE
   \FOR{$k = 1,2,\dots,|\mathcal{K}_{m+1}|$}
   \STATE $\text{Given } \overline{\mathcal{A}}^{\mathcal{K}_{m}} \text{, select top } L \text{ score points for cluster } \mathcal{K}^{(m+1)}_{k}$ 
   \STATE $\quad \triangleright \quad \phi \leftarrow (\phi(1),\phi(2), \dots, \phi(L)) \in \mathcal{P}^{\lbrace1,2 \dots ,|\mathcal{K}_{m}|\rbrace}_{L}$ 
   \STATE $\quad \triangleright \quad {Rank}\ (\mathcal{V}^{(m)}_{\phi(1)} \rightarrow \mathcal{K}^{(m+1)}_{k}) \geqslant \dots \geqslant {Rank}\ (\mathcal{V}^{(m)}_{\phi(L)} \rightarrow \mathcal{K}^{(m+1)}_{k})$
   \STATE $\quad \triangleright \quad \overrightarrow{\mathcal{N}}(l,i) \leftarrow \mathcal{F}^{(m)}_{\phi(l),i} \quad l = 1,2,\dots,L \quad i = 1,2,\dots,d_{IN} \quad \triangleright \text{ where }  \overrightarrow{\mathcal{N}} := \overrightarrow{\mathcal{N}}^{(m+1)}_{k}$
   \STATE $\text{Compute gates' activations } \sigma : \mathbb {R} \to (0,1) \text{ on top scores}$
   \STATE $\quad \triangleright \quad \sigma_{k,l} = \sigma(\alpha \ {Rank}\ (\mathcal{V}^{(m)}_{\phi(l)} \rightarrow \mathcal{K}^{(m+1)}_{k})  + \beta) \quad l = 1,2 ,\dots,L$
   \FOR{$j = 1,2,\dots,d_{OUT} \ $}
   \STATE $\mathcal{F}^{(m+1)}_{k,j} = \sum_{i=1}^{d_{IN}} \sum _{l=1}^{L} W_{l,i,j} \ (\sigma_{k,l} \cdot \overrightarrow{\mathcal{N}}(l,i)) + b_j $ 
   \ENDFOR
   \ENDFOR
   \STATE {\bfseries Return: $\mathcal{F}^{(m+1)}$}
    \end{algorithmic}
\end{algorithm}
\end{minipage}
\end{figure*}
\subsection{Computational Complexity}
Given an input affinity matrix $\mathcal{A}^{\mathcal{K}_{m}}$ and the desired number of output clusters $|\mathcal{K}_{m+1}|$, the cluster step has complexity equivalent to $\mathcal{O}(|\mathcal{K}_{m}|^{2}|\mathcal{K}_{m+1}|+|\mathcal{K}_{m+1}|^{2}|\mathcal{K}_{m+1}|)$, while the filter step has complexity $\mathcal{O}(|\mathcal{K}_{m}|^{2}|\mathcal{K}_{m+1}|)$. This analysis has been done without taking into account various possible optimizations, regarding for example multiplications in presence of sparse matrices. Moreover, the fastest schema consists in constructing the cluster hierarchy before the training process begins, then caching all intermediate $\mathcal{A}^{\mathcal{K}_{m}}$ and $\overrightarrow{\mathcal{N}}^{(m+1)}$. This way, the computational complexity for both steps is reduced to $\mathcal{O}(|\mathcal{K}_{m+1}| \times L \times d_{IN} \times d_{OUT})$. Such complexity constitutes an improvement w.r.t. the one deriving from Chebyshev~\cite{defferrard2016cnn} $\mathcal{O}(\mathcal{D}_{AVG} \times |\mathcal{A}_{m}| \times L \times d_{IN} \times d_{OUT})$ (where we indicate with $\mathcal{D}_{AVG}$ the average degree in $\mathcal{A}_{m}$), since $\mathcal{D}_{AVG} \times |\mathcal{A}_{m}| \approx |\mathcal{E}| > |\mathcal{A}_{m}| > |\mathcal{K}_{m+1}|$. Differently, the same conclusion cannot be easily achieved comparing our solution to GCN~\cite{kipf2017semi}, since its cost is $\mathcal{O}( |\mathcal{A}_{m}| \times d_{IN} \times (d_{OUT}+\mathcal{D}_{AVG}))$. 
\subsection{Model Analysis}
\begin{table}[b]
\centering
\caption{Impact of different graph's definitions on CIFAR-10 and Cross Subject NTU RGB+D, in terms of test set accuracy.}
\label{table:random_graph}
\resizebox{0.8\columnwidth}{!}{
\begin{tabular}{l c c c c}
\Xhline{2\arrayrulewidth}
\multicolumn{1}{c}{\textbf{Graph}} & \textbf{CIFAR} & \textbf{NTU-CS} \\ [0.05em]\hline \\[-0.8em]
Random & $69.0$ & $76.6$ \\
\textbf{Hand-crafted} & \textbf{$84.4$} & \textbf{$80.1$} \\
\Xhline{2\arrayrulewidth}
\end{tabular}}
\end{table}
\nuovo{\paragraph{Impact of the input graph} Regarding our proposal, how important is the quality and the integrity of the underlying graph? In terms of capabilities generalisation, what happens if we keep the signals unchanged and train our architecture on a random graph? Aiming to answer such questions, we conduct experiments replacing the designed shared graph (e.g. for CIFAR-10 the graph representing the 8-connectivity between pixels) with a random one, guaranteed to be connected and characterised by the same number of nodes and edges.}
\nuovo{As shown in Tab. \ref{table:random_graph}, we experience a considerable drop in performance when employing random connections between nodes (especially in the euclidean domain), consistently with what observed in~\cite{defferrard2016cnn}. Trivially, for graph convolutional neural networks, the compliance of the input affinity matrix with the domain-specific intrinsic bonds constitutes a crucial term for extracting meaningful features and, consequently, obtaining good level of accuracy on new and unseen data. 
In this respect, two ways may be investigated to improve such kind of approaches. On the one hand, drawing from the design principles underpinning kernel methods, many efforts may be put in designing better affinity measures between nodes. On the other hand, future works should move in a different direction, in which the affinity matrix is learned directly from the data, in a semi-supervised or completely unsupervised manner.}
\nuovo{\paragraph{Adaptation to heterogeneous graphs} Despite our model being originally conceived to assess a slightly different problem, we conjecture the possibility to extend it for heterogeneous graph classification, in which each example features a different affinity matrix. Indeed, a potential solution would be replacing the $U$ matrix in Eq.~7 (in the main paper) with the output of an auxiliary graph convolutional module, responsible for cluster memberships computation (given node features and the affinity matrix). As a consequence, affinities would completely depend on learned vertex representations.} 
\subsection{Limitations}
Since the computational complexity is approximately quadratic with the number of nodes, it is difficult to scale our method to very large graphs (e.g. $> 10^5$ nodes), in terms of both time complexity and memory footprint. For this reason, future works should investigate strategies that avoid expensive computations implied by manipulations of the affinity matrix.
Further, several experiments did not show a considerable benefit (in terms of accuracy improvements) from learning the cluster hierarchy through the use of information coming from the task supervision. This aspect, which may be due to some sort of lack in the current implementation, lead the authors to two observations. Firstly, since labels seem not to encourage preferable directions in the clusters' loss landscape, we observed slight improvements in a fully end-to-end training, with respect to a two-step optimization of the cluster hierarchy ($\mathcal{L}_K$) and the feature extractors ($\mathcal{L}_0$). Secondly, future studies should look into this matter in greater depth, trying to understand for which kind of problems a learnable routing on the underlying graph could provide considerable improvements against traditional architectures.
\end{document}